\documentclass[11pt]{article}
\usepackage{eamt18}
\usepackage{times}
\usepackage{url}
\usepackage{latexsym}
\usepackage{multirow}
\usepackage[small,bf]{caption} 
\usepackage[english]{babel}
\usepackage[utf8x]{inputenc}
\usepackage{amsmath}
\usepackage{graphicx}
\usepackage{natbib}
\usepackage{float}

\usepackage[justification=justified]{caption}

\usepackage[usenames, dvipsnames]{color}
\setcitestyle{square}

\RequirePackage{marginnote}
\makeatletter
\let\oldmarginnote\marginnote
\renewcommand*{\marginnote}[1]{%
   \begingroup%
   \ifodd\value{page}
     \if@firstcolumn\reversemarginpar\fi
   \else
     \if@firstcolumn\else\reversemarginpar\fi
   \fi
   \oldmarginnote{#1}%
   \endgroup%
}
\makeatother

\begin{document}

\title{Investigating Backtranslation in Neural Machine Translation}


\author{
\begin{tabular}{c}
\shortstack{Alberto Poncelas, Dimitar Shterionov, Andy Way,}\\
\shortstack{Gideon Maillette de Buy Wenniger \and Peyman Passban}\\
\end{tabular}
\\
School of Computing, DCU, ADAPT Centre\\
{\tt \{firstname.lastname\}@adaptcentre.ie}
}

\date{}

\maketitle

\begin{abstract}
A prerequisite for training corpus-based machine translation (MT) systems -- either Statistical MT (SMT) or Neural MT (NMT) -- is the availability of high-quality parallel data. This is arguably more important today than ever before, as NMT has been shown in many studies to outperform SMT, but mostly when large parallel corpora are available; in cases where data is limited, SMT can still outperform  NMT. 

Recently researchers have shown that back-translating monolingual data can be used to create synthetic parallel corpora, which in turn can be used in combination with authentic parallel data to train a high-quality NMT system. Given that large collections of new parallel text become available only quite rarely, backtranslation has become the norm when building state-of-the-art NMT systems, especially in resource-poor scenarios. 

However, we assert that there are many unknown factors regarding the actual effects of back-translated data on the translation capabilities of an NMT model. Accordingly, in this work we investigate how using back-translated data as a training corpus -- both as a separate standalone dataset as well as combined with human-generated parallel data -- affects the performance of an NMT model. We use incrementally larger amounts of back-translated data to train a range of NMT systems for German-to-English, and analyse the resulting translation performance. 
\end{abstract}

\section{Introduction}
Neural Machine Translation (NMT)~\citep{Cho2014,Sutskever2014,Bahdanau2014} is a relatively new machine translation (MT) paradigm that has quickly become dominant in both academic and industry MT communities, achieving state-of-the-art results~\citep{Bentivogli2016,Bojar2016,Junczys2016,Google2016,Castilho2017,Shterionov2017} on a range of language pairs and domains. As a corpus-based paradigm, the translation quality strongly depends on the quality and quantity of the training data provided. In comparison to statistical machine translation (SMT)~\citep{Koehn2010}, NMT typically requires more data to build a system with good translation performance~\citep{Koehn2017}. 

In many use-cases, however, the amount of good-quality parallel data available is insufficient to reach the translation standard required. In such cases, it has become the norm to resort to back-translating freely available monolingual data \citep{sennrich2015improving,belinkov2017synthetic,Domhan2017} to create an additional synthetic parallel corpus~\citep{sennrich2015improving} for training an NMT model. 

In this paper, we assert that this scenario has become the default in NMT without proper consideration of the merits of the approach. For example, \citet{Microsoft} present an algorithm for filtering noisy content from Web-scraped parallel corpora, in order to mitigate the ``pollut[ion] [of the Web] with increasing amounts of machine-translated content''. They note that their algorithm ``is capable of identifying machine-translated content in parallel corpora for a variety of language pairs, and that in some cases it can be very effective in improving the quality of an MT system ... thus challenging the conventional wisdom in natural language processing that `more data is better data'". Note too that \citet{Somers} demonstrates backtranslation (or `round trip’ translation) to be an untrusted means of MT evaluation. In the same vein, \citet{Way2013} notes that in order to show that MT is error-prone, ``sites like Translation Party ({\small\url{http://www.translationparty.com/}}) have been set up to demonstrate that continuous use of `back translation' – that is, start with (say) an English sentence, translate it into (say) French, translate that output back into English, ad nauseum – ends up with a string that differs markedly from that which you started out with". 

Surely, then, no-one would argue that building an MT system -- whether it be SMT or NMT -- with {\em solely} synthetic data is a good idea; after all, the premise underpinning the paper by \citet{Microsoft} was that adding machine-translated data to high-quality human-translated training data {\em harms} performance. Nonetheless, NMT developers have been seduced into using back-translated data as a means of necessity; there is simply not enough authentic human-translated parallel data available to obtain high-quality results in all scenarios where we would like to deploy NMT. Somewhat surprisingly, despite the inherent problems noted above, adding back-translated data does help improve the quality of NMT output!

In this paper we set out to systematically test from the ground up the merits of back-translated data. We investigate three scenarios: (i) NMT systems trained on `perfect' human-translated (authentic) data; (ii) using only back-translated (synthetic) data for training NMT systems; and (iii) NMT systems trained on a combination of human-translated and back-translated data. We systematically create multiple training corpora of increasing sizes, using training sets with authentic, synthetic and hybrid (authentic $+$ synthetic) data. 

For the hybrid case we  increment the back-translated to human-generated data ratio and observe the quality of the resulting NMT systems. We aim to identify to what extent adding synthetic data improves (or harms) the translation capabilities of NMT systems. That is, we investigate whether backtranslation as a core technique in NMT has any limits; given that synthetic data is generated via another imperfect MT system, we hypothesise that NMT trained with `imperfect' data will -- at some point -- undo any benefits from the `perfect' (human-translated) data, and lead the NMT to degrade in performance.\footnote{Note that this should not be confused with the problem of overfitting, where the NMT system learns the training data very well but fails to generalize, with the result that it performs poorly on unseen data. 
} 

In all our experiments, we exploit data that is widely used in the academic community for researching the quality of MT. The datasets that we use in our experiments all come from the Translation Task of the Tenth Workshop on Machine Translation in 2015 (WMT 2015 \citep{Bojar2015}).\footnote{\url{http://www.statmt.org/wmt15/}} To build our NMT systems we use OpenNMT-py (the pytorch port of OpenNMT \citep{opennmt}) with standard settings that allows for easy replicability of our experiments. 

The remainder of the paper is structured as follows: Section~\ref{sec:background} presents related work on using back-translated and other synthetic data in MT. Section~\ref{sec:effects} explains how back-translated data affects the training and quality of an NMT system. Our data is described in Section~\ref{sec:data}, and our experiments are outlined in Section~\ref{sec:experiments}. The results are summarised and analysed in Section~\ref{sec:results}. We conclude in Section~\ref{sec:conclusion} with final remarks and future work plans.

\section{Related Work}\label{sec:background}
Recent studies have shown different approaches to exploiting monolingual data to improve NMT. \citet{Glehre2015} present two approaches to integrate a language model trained on monolingual data into the decoder of an NMT system. Similarly, \citet{Domhan2017} focus on improving the decoder with monolingual data. While these studies show improved overall translation quality, they require changing the underlying neural network architecture. In contrast, backtranslation allows one to generate a parallel corpus that, consecutively, can be used for training in a standard NMT implementation as presented by \citet{sennrich2015improving}. \citet{sennrich2015improving} use 4.4M sentence pairs of authentic human-translated parallel data to train a baseline English $\rightarrow$ German NMT system that is later used to translate 3.6M German and  4.2M English target-side sentences. These are then mixed with the initial data to create human $+$ synthetic parallel corpora which are then used to train new models. Due to the good results that were obtained, adding synthetic data has become a popular step in the NMT training pipeline~\citep{sennrich2016edinburgh,di2017fbk,lo2017nrc}.

\citet{Karakanta2017} use back-translated data to improve MT for a low-resource language, namely Belarusian (BE). They transliterate a high-resource language (Russian, RU) into their low-resource language (BE) and train a BE$\rightarrow$EN system, which is then used to translate monolingual BE data into EN. Finally, an EN$\rightarrow$BE system is trained with that back-translated data.

The work of \citet{park2017building} presents an analysis of models trained only with synthetic data. In their work, they train NMT models with parallel corpora composed of: (i) synthetic data in the source-side only; (ii) synthetic data in the target-side only; and (iii) a mixture of parallel sentences of which either the source-side or the target-side is synthetic. 

Note too that in contrast to the efforts of \citet{Microsoft}, backtranslation has been applied successfully in PBSMT. \citet{Bojar2011} use back-translated data to optimize the translation model of a PBSMT system and show improvements in the overall translation quality for 8 language pairs. 

\section{Issues involved in creating back-translated parallel data}\label{sec:effects}

Intuitively, MT models built using synthetic data should not perform well. A text translated by a machine can contain errors, so a model trained on such data may learn and replicate these mistakes. While \citet{sennrich2015improving} demonstrated that using back-translated data (in combination with human-translated data) during training can have a positive impact on the performance of the model, we hypothesize that the performance of the model will degrade if the synthetic data is overly dominant in the training set, i.e. the benefit of using high-quality authentic parallel data may be outweighed by the synthetic back-translated data.

We investigate our hypothesis through a systematic analysis of NMT models trained on different-sized parallel datasets containing increasing amounts of back-translated data. While we acknowledge the plethora of factors (e.g. vocabulary size, word-segmentation, learning optimizer, learning rate, total amount of training steps/minibatches etc.) that may impact such an analysis, with this work we aim to provide a solid experimental baseline NMT set-up using freely available data. This baseline builds a clearer picture about the progressive effects of adding synthetic data to the training corpus of NMT engines. To the best of our knowledge, such an analysis has not been performed at the time of writing.

Furthermore, we compare NMT systems built on authentic-only data to systems built on synthetic-only data and put the two extremes to a test. We hypothesise that only synthetic data will not be enough to train an NMT system with good performance due to the errors mediated by the initial MT system used to generate that data. However, our results are more than a little surprising. We present detailed analysis of our empirical results in Section~\ref{sec:results}.


\section{Data}\label{sec:data}
For the scope of this work, we use the German--English parallel data of the WMT 2015 Translation task \citep{Bojar2015}. This corpus is shuffled, tokenized, truecased and cleaned (removing sentences of length over 126 words). In total, it contains 4.48M sentence pairs (225M words).

In order to explore the effects of back-translated data, we use human-translated (authentic) and back-translated (synthetic) data in three possible configurations:
\begin{itemize}
    \item Authentic data only: Models are trained  using authentic data only. Such models provide a baseline that any other model can be compared to. This is the baseline scenario for quality of data.  Furthermore, such models represent a use-case where an industry partner supplies authentic data to MT engineers in order to build an NMT system.
    \item Synthetic data Only: Models are built using  back-translated data only. Such models represent the case where no parallel data is available but monolingual data can be translated via an existing MT system and provided as a training corpus to a new NMT system. Such cases appear as the other extreme, or the worst-case scenario for quality of data. They reflect resource limitations, either due to the physical unavailability of data, i.e. low-resource languages, or due to economic reasons. Using  synthetic data only might also be an option in cases where a high-quality model trained on real data is available, but the translation task is on a very different domain than the training data. In this case using the high-quality model to back-translate domain-specific monolingual target data, and then building a new model with this  synthetic training data, might be useful for domain adaptation.
    \item Hybrid data: Models are built using a base dataset of 1M authentic sentence pairs combined with differing amounts of back-translated data. This is the most interesting scenario (similar to~\citet{sennrich2015improving}) which allows us to trace the changes in quality with increases in synthetic-to-authentic data ratio.
\end{itemize}


All the models that we built are evaluated using the same test set. This test set is provided by WMT 2015 news translation task. It consist of a collection of 2169 sentences from the news domain. These sentences have also been tokenized and truecased.

\section{Experimental set-up}\label{sec:experiments}
We train sequence-to-sequence NMT models~\citep{Sutskever2014} based on recurrent neural networks with an attention mechanism~\citep{Bahdanau2014,luong-pham-manning:2015:EMNLP}. The NMT framework we use is OpenNMT~\citep{opennmt} and in particular its pytorch\footnote{\url{http://pytorch.org}} port.

Our set-up follows the OpenNMT guidelines,\footnote{\url{http://opennmt.net/Models/}} that indicate that the default training configuration is reasonable for training a German-to-English model on WMT 2015 data.

We acknowledge the multitude of parameters and values that one can tweak in the set-up of an NMT system, leading to systems with significantly different performance. Moreoever, the choice of these parameters often depends on the training data. In our experiments, however, we have focused on a static NMT set-up, where the different parameters (e.g. the NMT learning optimizer, number of epochs, etc.) are common for all systems we train. The decision on our set-up is based on two factors: (i) by limiting the variability of parameters, we can more easily investigate the effects of back-translated data by directly comparing the translation quality of the resulting NMT systems; and (ii) while certain new architectures such as  \emph{Transformer}~\citep{NIPS20177181} or different settings might obtain even better results, our goal here is not to build the absolutely best possible systems, but rather use configurations that are representative of what is used in the field and allow easy replication. 

Specifically, we use a 2-layer LSTM \citep{hochreiter1997long} with 500 hidden units, a vocabulary size of 50,002 for the source language and 50,004 for the target language. The model is trained for 13 epochs, using the stochastic gradient descent learning optimizer and a batch size of 64. Any unknown words in the translation are replaced with the word in the source language that has the highest attention.

We first trained a baseline $DE \rightarrow EN$ model on $1,000,000$ parallel sentences of authentic data (\textit{base dataset}) and a baseline $EN \rightarrow DE$ model on the same data set with source and target sides swapped around. The latter model is used for backtranslation to create \textit{synthetic dataset}s. We found that using 1M sentences to train the model was sufficient for `good enough' translations. To determine this, we  performed preliminary tests that involve human evaluation alongside automatic metrics (on a random sample of the outputs) with models trained on other data sizes.\footnote{These experiments go beyond the scope of this work and are not included in the current paper.} When performing backtranslation, we also replace any unknown words with the word in English (the source language when performing the backtranslation) having the highest attention. We used this engine to then back-translate different portions of our original data set that we then used as parallel training data in two different scenarios: (i) by itself, i.e. synthetic data only, and (ii) in combination with the authentic data used to train the first engine, i.e. the hybrid models, as defined in Section~\ref{sec:data}. 

In order to make our comparison fair, we defined two cases of authentic data. The first one starts with the first 1,000,000 sentences and grows incrementally (adding 500,000 parallel sentences each time) until it contains 3,500,000 sentences, i.e. ranging between the 1$^{st}$ and the 3,500,000$^{th}$ sentence. We denote these sets as \emph{auth$_{0+}$}. The \emph{hybr} data sets are composed of the 1$^{st}$ 1,000,000 authentic sentences, combined with back-translated data for each following subset of 500,000 sentences. 

In the second case, the authentic data sets start from the 1,000,000$^{th}$ sentence. The first one contains 1,000,000 sentences; the next ones increment with 500,000 additional authentic sentences with the last one ranging between the 1,000,000$^{th}$ to the 4,480,000$^{th}$ sentence. These sets we refer to as \emph{auth$_{1+}$}. The \emph{synth} data sets are simply the back-translated data sets from the \emph{auth$_{1+}$} category.

In this way we compare engines trained on exactly the same original data -- \emph{auth$_{0+}$} to \emph{hybr} and \emph{auth$_{1+}$} to \emph{synth} -- which in one case has been partially or fully back-translated. 

In Table~\ref{table:backtranslated_coverage} we present the percentage of tokens (words, numbers and other symbols) of the test set that are covered by the vocabularies we use to build our models.


\begin{table}[!htbp]
\centering
\begin{center}
\begin{tabular}{  |p{0.8cm}|p{1.2cm}|p{1.2cm}||p{1.2cm}|p{1.2cm}|}
\hline
data size & \emph{auth$_{0+}$} & \emph{hybr} & \emph{auth$_{1+}$} & \emph{synthetic} \\
\hline	
1M	&	67.03\%	&	-	& 66.35\%	&	60.81\% \\
1.5M	&	67.15\%	&	66.14\%	 &66.44\%	&	60.93\% \\
2M	&	67.11\%	&	65.10\%	& 66.41\%	&	60.97\% \\
2.5M	&	67.25\%	&	64.60\%	& 66.36\%	&	61.03\% \\
3M	&	67.30\%	&	64.15\%	& 66.47\%	&	60.98\% \\
3.5M & 67.25\% & 63.77\% & 66.55\%	&	61.01\% \\
\hline
\end{tabular}
\caption{Coverage of the vocabularies on the tokens in the test set. \label{table:backtranslated_coverage} }
\end{center}
\end{table}

\section{Results}\label{sec:results}

Tables~\ref{table:results_backtranslated_hybrid} and \ref{table:results_backtranslated_synth} show the evaluation scores of the models we trained for the authentic-to-hybrid and authentic-to-synthetic cases, respectively. We use a number of common evaluation metrics -- BLEU \citep{papineni2002bleu}, TER \citep{snover2006study}, METEOR \citep{banerjee2005meteor}, and {\small CHR}F \citep{popovic2015chrf} -- to give a more comprehensive estimation of the comparative translation quality. With the exception of TER, the higher the score, the better the translation is estimated to be; for TER, being an error metric, the lower the score, the better the quality.

\begin{table}[!htbp]
\centering
\begin{tabular}{|p{0.35cm}|p{1.7cm}|p{1.7cm}|p{2cm}|}
\hline
\multirow{6}{*}{\rotatebox[origin=c]{90}{\centering 1M lines}} & & 1M auth.  & - \\
\hline
& BLEU	&	0.2278	&	-	\\
& TER$\downarrow$ 	&	0.5748	&	-	\\
& METEOR	&	0.269	&	- \\
& {\small CHR}F1	&	48.7336	&	-	\\	
\hline
\hline
\multirow{8}{*}{\rotatebox[origin=c]{90}{\centering 1.5M lines}} & & 1.5M auth.  & 1M auth. + 0.5M synth. \\					
\hline	
& BLEU$\uparrow$	&	0.2347	&	0.2378	\\
& TER$\downarrow$ 	&	0.5702	&	0.5681	\\
& METEOR$\uparrow$	&	0.2735	&	0.2751	\\
& {\small CHR}F1$\uparrow$	&	49.2973	&	49.5145	\\
\hline
\hline					
\multirow{8}{*}{\rotatebox[origin=c]{90}{\centering 2M lines}} & & 2M auth.  & 1M auth. +1M synth.  \\					
\hline					
& BLEU$\uparrow$	&	0.2382	&	0.2421	\\
& TER$\downarrow$ 	&	0.5646	&	0.5644	\\
& METEOR$\uparrow$	&	0.2755	&	0.2771	\\
&{\small CHR}F1$\uparrow$	&	49.6164	&	49.6818	\\
\hline					
\hline					
\multirow{8}{*}{\rotatebox[origin=c]{90}{\centering 2.5M lines}} & & 2.5M auth.  & 1M auth. + 1.5M synth.  \\					
\hline					
& BLEU$\uparrow$	&	0.2419	&	0.242	\\
& TER$\downarrow$ 	&	0.5592	&	0.5622	\\
& METEOR$\uparrow$	&	0.2786	&	0.2784	\\
& {\small CHR}F1$\uparrow$	&	50.015	&	49.8781	\\
\hline					
\hline					
\multirow{8}{*}{\rotatebox[origin=c]{90}{\centering 3M lines}} & & 3M auth.  & 1M auth. + 2M synth.  \\					
\hline					
& BLEU$\uparrow$	&	0.2446	&	0.2442	\\
& TER$\downarrow$ 	&	0.5572	&	0.5621	\\
& METEOR$\uparrow$	&	0.2792	&	0.2785	\\
& {\small CHR}F1$\uparrow$	&	50.1999	&	49.9244	\\
\hline		
\hline					
\multirow{8}{*}{\rotatebox[origin=c]{90}{\centering 3.5M lines}} & & 3.5M auth.  & 1M auth. + 2.5M synth.  \\					
\hline					
& BLEU$\uparrow$	&	0.2435	&	0.2413	\\
& TER$\downarrow$	&	0.5586	&	0.5651	\\
& METEOR$\uparrow$	&	0.2788	&	0.277	\\
& {\small CHR}F1$\uparrow$	&	50.0785	&	49.584	\\
\hline
\end{tabular}
\caption{Results of models using human-translated or authentic data and back-translated or synthetic data from the \emph{auth$_{0+}$} and \emph{hybr} sets.
\label{table:results_backtranslated_hybrid} }
\end{table}

\begin{figure*}[hbt]
\includegraphics[width=16cm, height=4cm]
{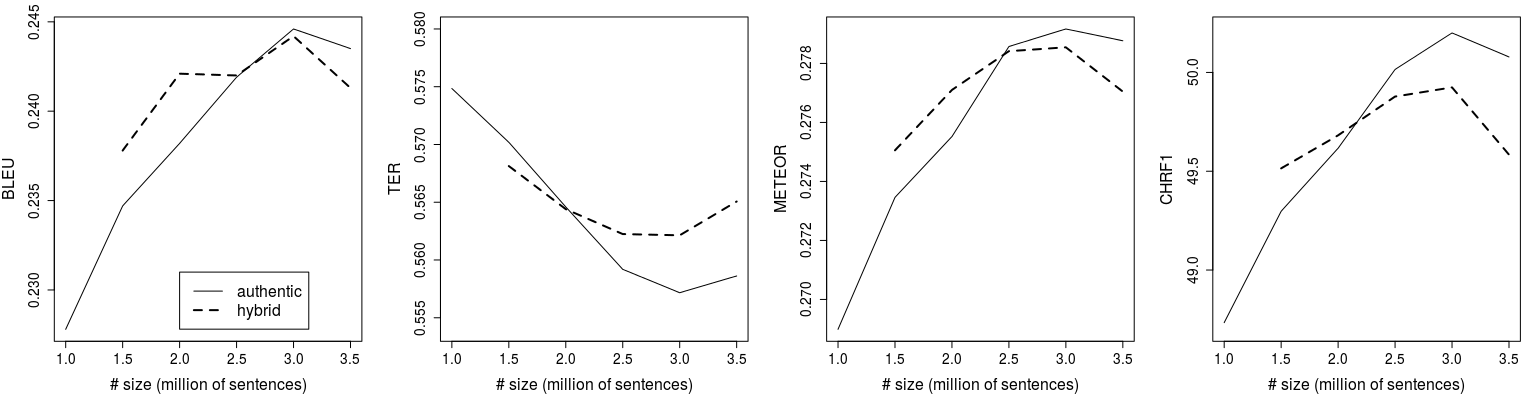}
\centering
\caption{Quality scores of NMT systems trained with different sizes of training data from the \emph{auth$_{0+}$} and \emph{hybr} sets. 
 \label{fig:auth_vs_hyb}
}
\end{figure*}

\begin{figure*}[hbt]
\includegraphics[width=16cm, height=4cm]
{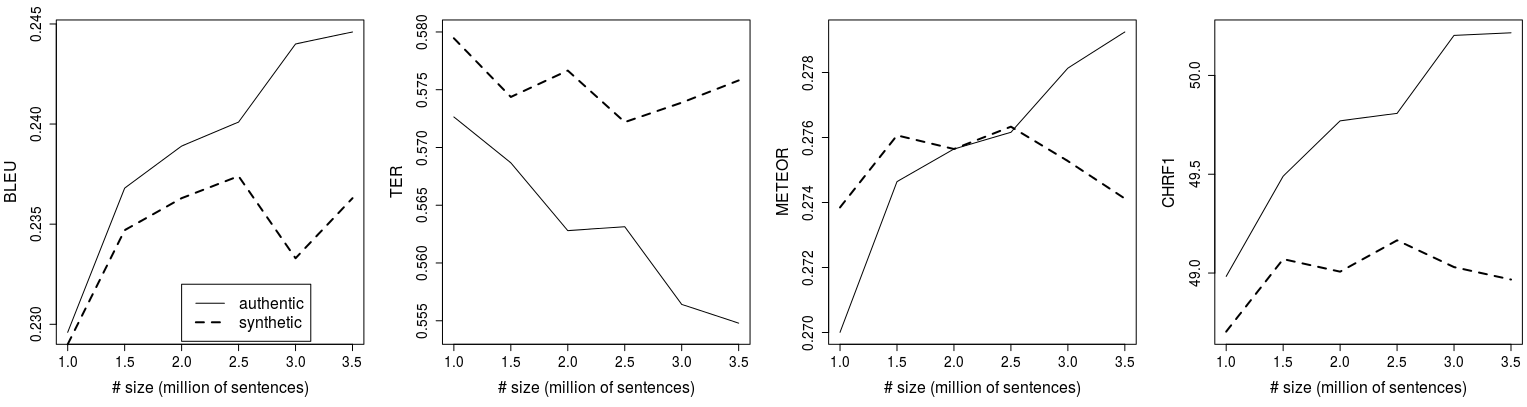}
\centering
\caption{Quality scores of NMT systems trained with different sizes of training data from the \emph{auth$_{1+}$} and \emph{synth} sets.
 \label{fig:auth_vs_synth}
}
\end{figure*}

In Figures~\ref{fig:auth_vs_synth} and \ref{fig:auth_vs_hyb} we illustrate how the BLEU and METEOR scores of our models (trained on authentic, synthetic and hybrid data) change with increases in the training data.

\begin{table}[!htbp]
\centering
\begin{tabular}{|p{0.35cm}|p{1.7cm}|p{1.7cm}|p{2cm}|}
\hline	
\multirow{6}{*}{\rotatebox[origin=c]{90}{\centering 1M lines}} &	& 1M auth.   & 1M synth. \\					
\hline					
& BLEU$\uparrow$	&	0.2296	&	0.229	\\
& TER$\downarrow$ 	&	0.5726	&	0.5795	\\
& METEOR$\uparrow$	&	0.2700	&	0.2738	\\
& {\small CHR}F1$\uparrow$	&	48.9829	&	48.7035	\\
\hline					
\hline					
\multirow{6}{*}{\rotatebox[origin=c]{90}{\centering 1.5M lines}} & & 1.5M auth.  &  1.5M synth. \\					
\hline					
& BLEU$\uparrow$	&	0.2368	&	0.2347	\\
& TER$\downarrow$ 	&	0.5687	&	0.5744	\\
& METEOR$\uparrow$	&	0.2746	&	0.2761	\\
& {\small CHR}F1$\uparrow$	&	49.4900	&	49.0705	\\
\hline					
\hline					
\multirow{6}{*}{\rotatebox[origin=c]{90}{\centering 2M lines}} & & 2M auth.  &  2M synth. \\					
\hline					
& BLEU$\uparrow$	&	0.2389	&	0.2363	\\
& TER$\downarrow$ 	&	0.5628	&	0.5767	\\
& METEOR$\uparrow$	&	0.2756	&	0.2756	\\
&{\small CHR}F1$\uparrow$	&	49.7702	&	49.0069	\\
\hline					
\hline					
\multirow{6}{*}{\rotatebox[origin=c]{90}{\centering 2.5M lines}} & & 2.5M auth.  &  2.5M synth. \\					
\hline					
& BLEU$\uparrow$	&	0.2401	&	0.2374	\\
& TER$\downarrow$ 	&	0.5631	&	0.5722	\\
& METEOR$\uparrow$	&	0.2762	&	0.2763	\\
& {\small CHR}F1$\uparrow$	&	49.8079	&	49.1656	\\
\hline					
\hline					
\multirow{6}{*}{\rotatebox[origin=c]{90}{\centering 3M lines}} & & 3M auth.  &  3M synth. \\					
\hline					
& BLEU$\uparrow$	&	0.2440	&	0.2333	\\
& TER$\downarrow$ 	&	0.5564	&	0.5739	\\
& METEOR$\uparrow$	&	0.2781	&	0.2753	\\
& {\small CHR}F1$\uparrow$	&	50.2028	&	49.0301	\\
\hline					
\hline					
\multirow{6}{*}{\rotatebox[origin=c]{90}{\centering 3.5M lines}} & & 3.5M auth.  &  3.5M synth.$^{\ast}$ \\					
\hline					
& BLEU$\uparrow$	&	0.2446	&	0.2363	\\
& TER$\downarrow$	&	0.5548	&	0.5758	\\
& METEOR$\uparrow$	&	0.2792	&	0.2741	\\
& {\small CHR}F1$\uparrow$	&	50.2159	&	48.9671	\\
\hline
\end{tabular}
\caption{Results of models using human-translated or authentic data and back-translated or synthetic data from the \emph{auth$_{1+}$} and \emph{synth} sets.
\label{table:results_backtranslated_synth} }
\end{table}




\subsection{Authentic Data Models}
In Tables~\ref{table:results_backtranslated_hybrid} and \ref{table:results_backtranslated_synth}, we see that, as expected, building NMT systems with increasingly larger amounts of human-translated data improves performance: from a BLEU score of 0.2278 with 1M sentence pairs, to the best score of 0.2446 with 3M sentence pairs. This is an absolute improvement of 0.0168, or 7.4\% relative. We do, however, see a slight drop when we build our NMT system with 3.5M sentence pairs. All these findings are corroborated by the other five MT evaluation metrics.

\subsection{Synthetic Data Models}
Earlier in the paper, we suggested that no-one would set out to build an NMT system using solely synthetic data. However, our results show this to be far from the crazy idea it seemed at the outset (see Table~\ref{table:results_backtranslated_synth} and Figure~\ref{fig:auth_vs_synth}). Using 1M sentence pairs of synthetic-only data (the first of the \emph{synth} data sets), we obtain a BLEU score of 0.229, which continues to rise as we add more synthetic data, achieving the best BLEU score of 0.2363 with 3.5M sentence pairs. This is an absolute improvement of 0.0073, or 3.2\% relative. Looking at the other metrics, the picture is rather more mixed; TER, METEOR and {\small CHR}F follow a more steady tendency.

It is clear, however, that the difference between the quality of engines trained on synthetic and authentic data is rather small. Moreover, the authentic and synthetic data sets of 1,000,000 sentences result in engines where the latter one actually performs better in terms of METEOR. However, even if smaller models built using synthetic data only can perform very close to the level of authentic-only models, it does not appear to be scalable, as the differences in the quality metrics between the two types of engines increase with larger data sizes, i.e. if we look at Figure \ref{fig:auth_vs_synth}, the quality of the models trained with synthetic data have a relatively lower increase in quality when more back-translated sentences are added.

We see in column \emph{synth} of Table \ref{table:backtranslated_coverage} that the coverage of models built using synthetic data do not increase when  more data is added, as they all are around 61\%. This coverage is much lower than for authentic data models (\emph{auth$_{1+}$} column), which have a coverage of more than 66\% for all training sizes. 


We put this discrepancy in performance down to the limits of the knowledge encoded by the NMT system used for back-translation. In particular, the sentences on the source side are the output of that system, and so (i) the vocabulary of these source-side sentences is always restricted; and (ii) these sentences will contain errors mediated by the initial NMT system. Given enough data, it will reach a steady point and not improve further. We observe this in Figure~\ref{fig:auth_vs_synth}. We can thus conclude that an NMT system trained on synthetic-only data can learn very well the knowledge encoded by the original system used for back-translation, and can even exceed its quality.


\subsection{Hybrid Data Models}
According to the results summarised in Table~\ref{table:results_backtranslated_hybrid} and Figure~\ref{fig:auth_vs_hyb}, the benefits of adding back-translated data presented in \citet{sennrich2015improving} are maintained in our experiments. We see that the hybrid model where 0.5M synthetic sentences are added in the training data (i.e. \textit{1M auth + 0.5M synth} column in Table \ref{table:results_backtranslated_hybrid}) performs better than the model built with 1M human-translated sentences. In fact, the same-sized hybrid model also outperforms the authentic-only model built with 1.5M sentence pairs.

Adding more and more synthetic data to the training set of an NMT systems causes BLEU scores to rise, as expected, with the best combination comprising 3M sentence pairs (1M authentic and 2M synthetic sentence pairs), which achieves a BLEU score of 0.2442, 0.0066 points absolute better than the smallest hybrid model, a relative improvement of 2.8\%.

We see in column \textit{hybr} of Table \ref{table:backtranslated_coverage} that the coverage of the hybrid models is not as high as for those built with authentic data only, but in all cases they are higher than for the synthetic-only datasets. We observe that the bigger the data set, the lower the coverage is. We expect that as more synthetic data is added, the more its vocabulary starts to dominate, pushing out words that are more frequent in real parallel data, but less frequent in synthetic data. Accordingly, we expect the coverage of hybrid models to tend to converge to the values of the synthetic models.


Figure~\ref{fig:auth_vs_hyb} shows how the quality of the hybrid models increases the more synthetic data is added. For smaller models, the slopes of the hybrid and authentic models are similar. However, the slope becomes less steep for models trained with 2M sentences or more, as in hybrid datasets with 2M sentence pairs half of it contains synthetic data.
\newline\newline




It is worth mentioning that models trained on synthetic or on hybrid data outperform the authentic-only models in the lower-sized training data sets. This indicates that in low-resource scenarios it makes sense to exploit back-translation in order to achieve a better NMT system. However, with synthetic-only data, at a given point the performance of the NMT system plateaus, while in the case of hybrid data the quality starts degrading as the synthetic data overpowers the authentic. In our experimental set-up and data we reached this point at a synthetic-to-authentic ratio of 2:1. In the future we will conduct more experiments with different data, data sizes and language pairs, as well as network set-ups to see whether a true tipping point emerges.

We believe this finding will have positive consequences especially for resource-poor scenarios. In particular, we hypothesise that using any existing MT system (or a combination of systems) to translate monolingual data in order to build an NMT system for the intended language direction with that data is likely to result in translation quality similar to that of the initial MT system.

\section{Conclusion and Future Work}\label{sec:conclusion}

In this work we studied the performance of NMT German-to-English models when incrementally larger amounts of back-translated (or synthetic) data are used for training. We analysed hybrid NMT models built by adding back-translated data to an initial set of human-translated (or authentic) data, and showed that while translation performance tends to improve when larger amounts of synthetic data are added, performance appears to tail off when the balance is tipped too far in favour of the synthetic data; in our experiments we see a drop in performance of 1.2\% for the 3.5M hybrid model compared to the 3M hybrid one. We plan to extend these experiments further in our future work, in order to figure out whether there exists a genuine tipping point, i.e. a ratio between the amount of synthetic and authentic data at which the model will achieve optimal performance, and beyond which the more synthetic data added, the worse the NMT quality becomes.

We also built models using synthetic data alone. To our surprise, the performance is quite good; the synthetic-only baseline model achieved quality very close to that of the authentic-only engines. Astonishingly, the synthetic-only engine trained with 1M sentences performs better as scored by METEOR than the authentic-only engine trained on the same amount of data. 

We believe that our findings have important repercussions for resource-poor scenarios, especially where some prior engine -- not necessarily an NMT system -- exists for the reverse language direction, as this can be used to create arbitrarily large amounts of back-translated data for bootstrapping an NMT engine for the other language direction. We will investigate this further in ongoing work.

In other future work, we also want to explore the effect of adding artificial data to different language pairs and domains. We envisage the current research as the first contribution to an ongoing investigation of the true merits and limits of back-translation. It may well turn out that adding incrementally larger amounts of back-translated data is less harmful than we expect, but at least doing this from the ground up will hopefully result in a set of principles for NMT practitioners, rather than the rather haphazard state of affairs we see before us today.

\section*{Acknowledgements}
This research is supported by the ADAPT Centre for Digital Content Technology, funded under the SFI Research Centres Programme (Grant 13/RC/2106).

This project has received funding from the European Union’s Horizon 2020 research and innovation programme under the Marie Skłodowska-Curie grant agreement No 713567.

\bibliographystyle{plainnat}
\bibliography{eamt18}

\end{document}